\documentclass{article}
\pdfoutput=1

\usepackage{arxiv}
\usepackage[utf8]{inputenc} 
\usepackage[T1]{fontenc}    
\usepackage{hyperref}       
\ifpdf\hypersetup{%
    pdftitle = {A CNN-based methodology for breast cancer diagnosis using thermal images},
    pdfsubject = {Preprint},
    pdfkeywords = {Machine Learning, Breast Cancer, Thermography, Hyper-parameters Optimization, CNN},
    pdfauthor = {Juan Pablo Zuluaga Gomez},
    pdfcreator = {\LaTeX},
    }
\fi
\hypersetup{
    colorlinks=true,
    linkcolor=black,
    citecolor=black,
    filecolor=black,
    urlcolor=black,
}
\usepackage{url}            
\usepackage{booktabs}       
\usepackage{amsfonts}       
\usepackage{nicefrac}       
\usepackage{microtype}      
\usepackage{lipsum}
\usepackage{tabularx}
\usepackage{graphicx, multirow}
\usepackage{gensymb}        
\usepackage{amsmath}
\usepackage{algorithm}      
\usepackage{algpseudocode}

\makeatletter
\def\BState{\State\hskip-\ALG@thistlm}
\makeatother
\title{A CNN-based methodology for breast cancer diagnosis using thermal images}

\author{J. Zuluaga-Gomez\textsuperscript{ a, d, e, }\thanks{Corresponding author: Juan Pablo Zuluaga, ORCiD: 0000-0002-6947-2706. Prognostics \& Health Management Team, Femto-ST Sciences \& Technologies, Besançon Cedex, France, 25000. Email: juan.zuluaga@eu4m.eu}
 , Z. Al Masry\textsuperscript{ a}
, K. Benaggoune\textsuperscript{ b}
, S. Meraghni\textsuperscript{ c}
, N. Zerhouni\textsuperscript{ a}
\\ \\
\textsuperscript{a}FEMTO-ST institute, Univ. Bourgogne Franche-Comté, CNRS, ENSMM, Besançon, France\\ \textsuperscript{b}Laboratory of Automation and Production engineering, Batna University, Batna, Algeria\\
\textsuperscript{c}LINFI Laboratory, University of Biskra, Biskra, Algeria\\
\textsuperscript{d}Electrical Engineering Department, University of Oviedo, Gijon, Spain\\
\textsuperscript{e}Universidad Autonoma del Caribe, Barranquilla, Colombia\\}

\begin{document}
\maketitle

\textbf{Highlights:}
\begin{itemize}
    \item Efficiency and reliability for breast cancer diagnosis through thermography 
    \item CNNs performance enhancement with data augmentation techniques 
    \item Smaller and simpler CNNs architectures perform better than state-of-the-art CNNs
    \item Trade-off measurement between data augmentation and the database size  
\end{itemize}

\begin{abstract}
\textbf{\textit{MicroAbstract}: A recent study from GLOBOCAN disclosed that during 2018 two million women worldwide had been diagnosed from breast cancer. This study presents a computer-aided diagnosis system based on convolutional neural networks as an alternative diagnosis methodology for breast cancer diagnosis with thermal images. Experimental results showed that lower false-positives and false-negatives classification rates are obtained when data pre-processing and data augmentation techniques are implemented in these thermal images.}\\
\textbf{Background:} There are many types of breast cancer screening techniques such as, mammography, magnetic resonance imaging, ultrasound and blood sample tests, which require either, expensive devices or personal qualified. Currently, some countries still lack access to these main screening techniques due to economic, social or cultural issues. The objective of this study is to demonstrate that computer-aided diagnosis(CAD) systems based on convolutional neural networks (CNN) are faster, reliable and robust than other techniques. \textbf{Methods:} Despite the disadvantages of the traditional machine learning techniques with spatial information, CNNs stand as main techniques for pattern recognition on images -or thermal images-. We performed a study of the influence of data pre-processing, data augmentation and database size versus a proposed set of CNN models. Furthermore, we developed a CNN hyper-parameters fine-tuning optimization algorithm using a tree parzen estimator. \textbf{Results:} Among the 57 patients database, our CNN models obtained a higher accuracy (92\%) and F1-score (92\%) that outperforms several state-of-the-art architectures such as ResNet50, SeResNet50 and Inception. Also, we demonstrated that a CNN model that implements data-augmentation techniques reach identical performance metrics in comparison with a CNN that uses a database up to 50\% bigger. \textbf{Conclusion:} This study highlights the benefits of data augmentation and CNNs in thermal breast images. Also, it measures the influence of the database size in the performance of CNNs. \linebreak
\end{abstract}


\section{Introduction}
\label{sec:introduction}

Breast cancer is the most commonly diagnosed cancer in women worldwide; then, it has become significant public health. It was the first leading cause of cancer-linked death among women in 2018, reaching approximately 15\% of the total number of registered cases \cite{9}. The early detection of breast cancer is imperative to reduce the mortality and morbidity index \cite{1,2,3}. Some studies suggest that emerging economies have almost a double risk of cancer, where the mortality-to incidence ratio in developed countries is 0.20, but in less developed countries is almost twice, thus 0.37 \cite{4,7}. Other factors like socioeconomic \cite{4,5}, aging, unhealthy lifestyle \cite{5,6,7,8}, environmental issues, and growth of the population may perhaps lead to higher risks. In perspective, Li et al. \cite{30} prove the correlation between body weight, parity, number of births, and menopausal status concerning breast cancer. On the other hand, some countries keep multiple barriers for developing an effective breast cancer screening system, e.g., organizational, psychological, structural, socio-cultural, and religious \cite{10}. Physicians, self-examination, and imaging techniques can perform detection of abnormalities in the breast, but a biopsy is the only way to confirm whether there is cancer \cite{65_1}. Imaging techniques like mammography, ultrasound, and magnetic resonance imaging currently stand as the main techniques for early breast cancer screening. However, limitations such as x-rays, expensiveness, dense tissue during young age, false positives (FP), and false-negative (FN) rates encouraged researchers and institutions to research alternative techniques like thermography deeply. Contrary to other modalities, thermography is a non-invasive, non-inclusive, radiation-free, and low-cost technique \cite{SurveyJuan}. Thus, thermography could be used to diagnose early-stage breast cancer e.g., young women and in dense breast patients. 
Frequently, these novel techniques, such as thermography, are coupled with computer-aided diagnosis (CAD) systems. A CAD system is a computational tool or algorithm capable of identifying patterns in many types of data e.g., clinical 2D and 3D clinical databases; consequently, several research teams are measuring the impact of CAD systems in the diagnosis of breast cancer patients \cite{24}.

Thermography measure the temperature based on infrared radiation. In medicine, the skin's surface temperature gives health insights because, the radiance from human skin is an exponential function of the surface temperature, in other words, it is influenced by the level of blood perfusion in the tumor \cite{25}, e.g. Krawczykm et al. \cite{26} summarize that thermography is well suited to detect changes in blood perfusion that are led by inflammation, angiogenesis, benign and malignant tumor. In 1956 the M. D. Lawson \cite{36} recorded for the first time the skin’s heat energy using a “thermocouple” \cite{36}, then authors presented similar devices e.g. the Pyroscan \cite{38}. On the one hand, thermography has advantages over other techniques, in particular when the tumor is in an early-stage or in dense tissue \cite{28,29}. On the other hand, the thermography stands as a technique capable of overcoming the limitations of mammography such as x-rays, painfulness during the test, and not-permissible cost in some underdevelopment countries. Consequently, during the last decades, there is an increasing research focused on machine learning techniques (MLT) for breast cancer diagnosis using thermal images; some researchers focus their works on the localization and size of tumors in phantoms and simulated models; but others scientists have been focused on characteristics like breast quadrants, menstrual cycle and acquisition protocols. 

During the last years, promising results have been achieved in various medical imaging applications for breast cancer diagnosis \cite{29_1,29_2} with CNNs. As mentioned in Section \ref{sec:relatedWork}, we concluded that CNNs have not been used widely in the past for breast cancer diagnosis with thermography, maybe because the CNNs were not efficient as texture or statistical features, or perhaps because the computing load was too high. Nonetheless, during the last years, CNNs techniques stand as one of the main techniques for pattern recognition in images -or thermal images-. In this work, we develop a novel CNN-CAD methodology that targets the public breast thermography database called “DMR-IR” proposed by Marquez \cite{52} and Silva et al. \cite{53}. This CNN-based study has five main contributions listed as follows:

\begin{itemize}
    \item We created baseline CNN models to replicate the results obtained by most of the recent studies regarding the DMR-IR database. This allowed us to find a weak spot during the training process that other studies have not to tackle previously. Therefore, we propose a new unbiased methodology to reduce further the likely training overfitting.
    \item In order to compare our CNN performance, here, we present a benchmark comparison between state-of-the-art architectures like ResNet, SeResNet, VGG16, Inception, InceptionResNetV2, and Xception. We demonstrated that smaller and less-complex CNN architectures are much better for the DMR-IR database.
    \item Following some survey articles \cite{SurveyJuan,56}, we concluded that just a couple of authors had employed CNN techniques instead of texture and statistical features for this database. Thus, we propose a better CNN architecture than the state-of-the-art ones, but also limiting the overfitting during the training process.
    \item Besides the comparison between state-of-the-art and our proposed CNN architectures, we also developed a hyper-parameters optimization algorithm based on a tree parzen estimator to increase the performance of the models further.
    \item Finally, knowing that normally the databases in the biomedical environment are limited, expensive, hard to acquire and changes depending on the acquisition protocol, we measured the influence of data augmentation and database size during the training, with the intention to suggest a minimum number of patients to obtain an effective CAD system. 
\end{itemize}
Although the CAD system has been trained with the DMR-IR database, this approach is useful for other databases of thermal breast images. The outline of this article is as follows. Section 2 explains the related work and the main ideas behind thermography and the influence of breast tumors in temperature changes. Section 3 describes the acquisition protocol and the methodology for data pre-processing and data augmentation. In order to illustrate the novelty and advantage of our methodology regarding other studies, we compared the results of four sets of experiments in Section 4. Lastly, discussion and conclusions are presented in Section 5 and 6.

\section{Current techniques for breast cancer diagnosis from thermal images}
\label{sec:relatedWork}

The rapid growth of virtual collaboration, programming tools and computing performance have raised the interest of many researchers over CAD systems in the biomedical area. Arena et al. \cite{47} summarize the benefits of thermography over the classical methods for breast cancer diagnosis. They tested a weighted algorithm in 109 tissue proven cases of breast cancer, that generates positive or negative result based on six features (threshold, nipple, areola, global, asymmetry and hot spot). Krawczyk et al. \cite{26} propose an ensemble method for clustering and classification in breast cancer thermal images; additionally, a 5x2 K-fold cross-validation was made to reduce the bias and obtain a more robust model. Later, in 2009 Schaefer et al. \cite{54} performed a fuzzy logic classification algorithm on 150 cases having an accuracy of 80\%; they explain that statistical feature analysis is a key source of information for achieving high accuracy, i.e. symmetry between left and right breast, standard temperature deviation, max-min temperatures, among others. In addition, some researchers centered their studies on the tumor’s characteristics and behavior such as Partridge and Wrobel \cite{48}, whose designed a method using dual reciprocity joined with genetic algorithms to localize tumors, where they found that smaller and deeply located tumors produce only a limited thermal perturbation making harder their detection. Contrary, it is possible to determine the tumor’s characteristics when the thermal surface behavior is known, Das and Mishra \cite{49} affirmed this. Kennedy et al. \cite{50} make a comparison between breast cancer screening techniques such as thermography, mammography and ultrasound. 

\begin{figure}[htb]
\begin{center}
\includegraphics[trim=0cm 45cm 70cm 0cm, clip, width=3.5in]{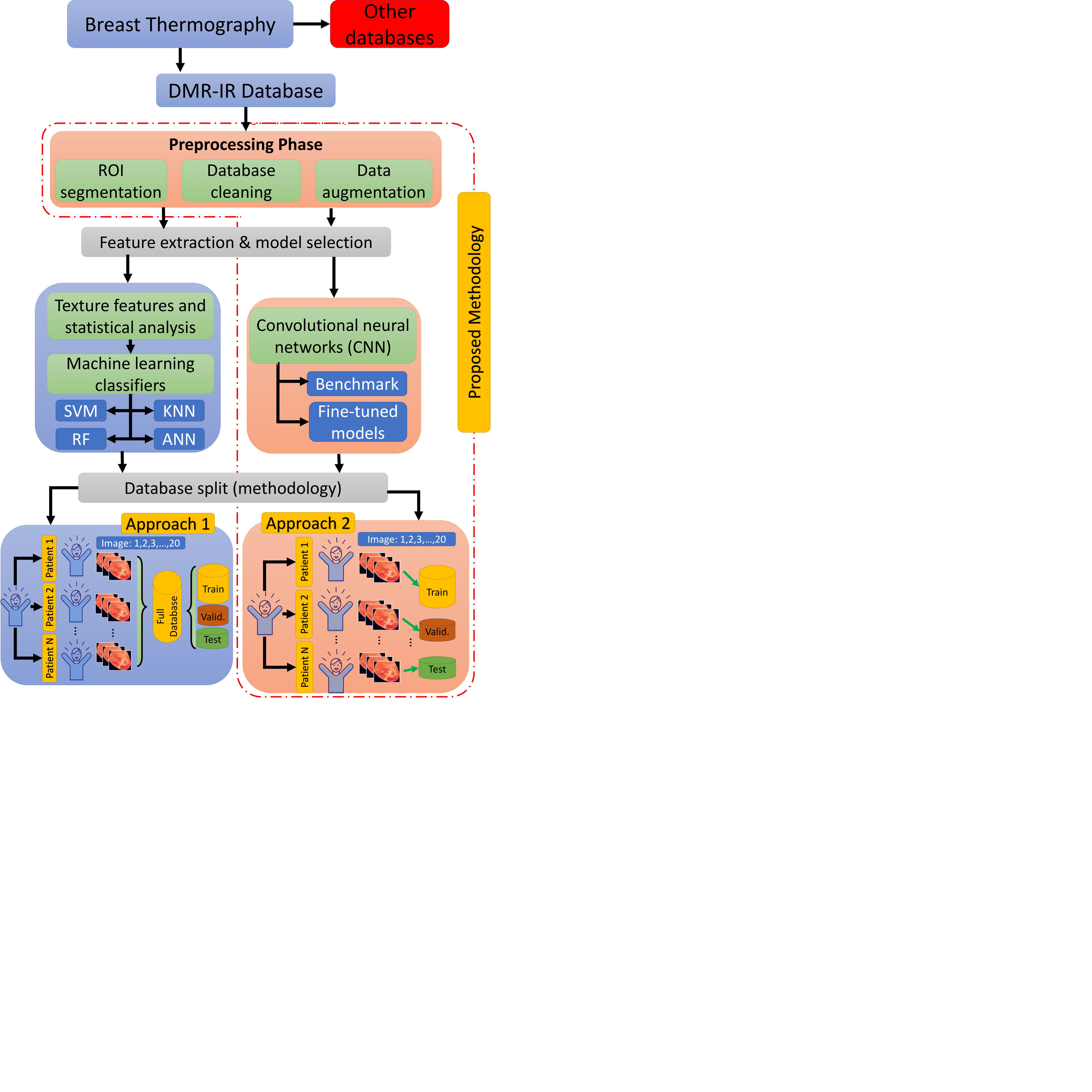}
\caption{Current approaches for feature extraction and database splitting in DMR-IR database.}
\label{methodology_1}
\end{center}
\end{figure}

The majority of studies related to CAD systems and infrared imaging techniques for breast cancer diagnosis employ the public and web-available database from \cite{52,53}. The database is composed of 1140 images from 57 patients, where 38 carried anomalies and 19 were healthy women from Brazil; additionally, each patient has a sequence-set of 20 images. The interpretation of a breast thermography test could be either, temperature matrices or heat map images, as it is the proposed database (also called DMR-IR database). Thermal images share similarities with a standard gray-scale or colored image; thus, mostly of the studies over the DMR-IR database try to identify texture and statistical features from those thermal images and matrices. Rajendra et al. \cite{51} built an algorithm using support vector machines for automatic classification of normal and malignant breast cancer. They extracted from the DMR-IR database texture features from the co-occurrence matrix and statistical features from the temperature matrix, achieving an accuracy, sensitivity and specificity of 88.1\%, 85.71\% and 90.48\%, respectively. Araujo \cite{55} presented a symbolic data analysis on 50 patients' thermograms obtaining four variables from the thermal matrices; also, he applied leave-one-out cross validation framework during the training process obtaining 85.7\% of sensitivity for the malignant class and accuracy of 84\%. Mambou et al. \cite{2} describe a method to use deep neural networks and support vector machines for breast cancer diagnoses, but also, they call attention to camera sensitivity and physical model of the breast. Nevertheless, in these above-mentioned studies are not stated how it is split the 1140-image database, neither if during the training process the whole twenty-image sequences from each patient belongs to either, train or test set or both datasets simultaneously.

Table \ref{tab:summary_studies} shows a summary of the last studies using the DMR-IR database. Their approaches have two big branches: on the one hand, a big portion of these studies have used texture and statistical features, where they have achieved 95\% accuracy \cite{60, 67}. On the other hand, other authors rather have chosen CNNs, achieving more than 90\% accuracy \cite{65,66}. A main concerning with Table \ref{tab:summary_studies} studies is that each one presents a variable number of patients, which allows us to infer that the database has suffered changes over the last years such as, the inclusion of new patients. Besides that, it is important to recall that most of these works do not mention sufficient information regarding the database split methodology during the training framework. Then, there are two possible approaches. On the one hand, it is possible to stack all twenty-image sequences from each patient in one database and then split it in train/test datasets. On the other hand, each patient’s image-sequence is assigned to either, train or test set as presented in Figure \ref{methodology_1}. Indeed, our main contributions are done under Approach 2 in the red-delimited are of Figure \ref{methodology_1}; also, this figure defines the pre-processing, training and database split frameworks of Table \ref{tab:summary_studies} studies.  

\begin{table}[htb]
\centering
\caption{Summary of algorithms based on machine learning techniques, statistical and texture features. These studies are based on the DMR-IR database for breast cancer diagnosis}
\label{tab:summary_studies}
\begin{tabular}{|ccp{8cm}p{2cm}p{2cm}|}
\toprule
    Ref. & Year & Machine learning technique / Extracted Features & Acquisition \newline protocol & Numb. of patients \newline (Malig./Heal.) \\
\midrule
\cite{51} & 2012 & Support vector machines (SVM) for texture features and statistical analysis & Static & 50 (25/25) \\
\cite{53} & 2014 & K-nearest neighbors (KNN) algorithm to classify Affine Scale-Invariant Feature Transform (database owners) & Static \& \newline Dynamic & 149 \\
\cite{55} & 2014 & They obtained the nterval data in the symbolic data analysis \& statistical analysis & Static & 50 (31/19) \\
\cite{58} & 2015 & Extended hidden Markov models for breast segmentation & Static & 160 \\
\cite{59} & 2015 & K-means and clustering from silhouette, Davies-Bouldin and Calinski-Harabasz indexes & Dynamic & 22 (11/11) \\
\cite{60} & 2016 & BayesNet, KNN \& Radom Forest (RF) models for pixel intensity and time series analysis \& Static & Dynamic & 80 (40/40) \\
\cite{61} & 2017 & SVM \& Genetic Algorithm (GA) for classification of normalized breast thermograms using local energy features & Static & 100 (47/53) \\
\cite{2} & 2018 & SVM, Artificial Neural Networks (ANN), Deep ANN, Recurrent ANN & Static & 64 (32/32) \\
\cite{62} & 2018 & SVM, KNN \& ANN for texture features and statistical analysis & Static & 80 (30/50) \\
\cite{63} & 2018 & Bilateral asymmetry and statistical analysis for annotation of thermograms & Static & 100 (49/51) \\
\cite{64} & 2018 & Multi-Layer Perceptron (MLP), DT \& RF using Zernike and Haralick moments as features & Static & 100 (30/70) \\
\cite{65} & 2018 & CNN models for static \& dynamical analysis & Static \& \newline Dynamic & 137 (42/95) \\
\cite{66} & 2019 & State-of-the-art benchmark of several CNN architectures & Static & 216 (41/175) \\
\cite{67} & 2019 & Learning-to-rank (LTR) and texture analysis methods like histogram of oriented gradients & Dynamic & 56 (37/19)\\
\bottomrule
\end{tabular}
\end{table}

During the last six years, several reviews concerning infrared technologies have emerged, well delimiting the status, main protocols and mew directions of breast cancer imaging diagnosis techniques \cite{12, 56, 57, SurveyJuan}. One significant fact mentioned in those reviews is that CAD thermography systems need to reduce the utmost non-relevant information in the thermal images. A thermal image typically has unnecessary areas such as chess, background and other parts of the body, but this data is not useful and acts as noise during the training in CNN models or during the features identification process. Hence, the process  provides a clean breast image without a non-necessary area to a CAD system is executed by a region of interest (ROI) algorithm. Regarding the DMR-IR database \cite{52,53}, several authors have based their research on ROI algorithms rather than identifying patterns in thermal images e.g. \cite{58} use extended hidden Markov models (EHMM), BayesNet and random forest in a 160-individuals for optimization of breast segmentation algorithms. Sathish et al. \cite{61} extracted the breast’s ROI and uses asymmetry and local energy features of wavelet sub-bands to determine whether the patient has cancer. They also concluded that the normalization of each thermal image could improve the general efficiency of the segmentation algorithm. In addition, extreme learning machines \cite{64} and efficient coding \cite{63} have been used for ROI segmentation.

To summarize, it is essential to recall that several aspects influence the overall performance and complexity of a given system such as pre-processing techniques, features extraction, statistical analysis, a region of interest selection, CAD technique (machine learning approach), training framework, database splitting and post-processing. Nevertheless, algorithm’s complexity is not directly proportional to the algorithm’s performance. This paper differs from previously published studies (see Table \ref{tab:summary_studies}) using the DMR-IR database \cite{52,53}, since our main goal is to demonstrate that a CNN-based CAD system outperforms the texture features based algorithms from Table \ref{tab:summary_studies}, but at the same time, it is less complex, easier to train and capable of generalizing more when new patients come. This paper (i) presents a new methodology (see Figure \ref{methodology_1}) which has a greater performance outperforming some studies from Table \ref{tab:summary_studies}; (ii) compares the performance of several state-of-the-art CNN architectures (benchmark); (iii) proposes a methodology for increase the CNN performance when hyper-parameters optimization is used and; (iv) determines the impact of data augmentation, data pre-processing and database size during and after the training process.

\section{Database description and proposed methodology}
\label{sec:methodology}

In this research, we propose several CNN-based experiments for the diagnosis of breast cancer using thermal images using the popular, free, and public available DMR-IR database, which is accessible through a user-friendly online interface (http://visual.ic.uff.br/dmi). In the first step, we applied data pre-processing and data augmentation for each thermal image e.g. crop, resizing, and breast normalization. In the second step, we defined several sets of interconnected experiments that tested different CNNs architectures under different training frameworks based on the database split methodologies from Figure \ref{methodology_1} (Section \ref{sec:relatedWork}) and following the Figure \ref{workflow} workflow. As our study is based on CNN, the overview of a training process is as follows: firstly, each thermal breast image is forwarded through a given number of hidden layers until a loss function is computed; secondly, the loss function is back-propagated into these layers, modifying the weights in accordance with an optimizer e.g. Adam. Finally, this procedure is looped for given N numbers of epochs until it reaches the desired performance metric value. 

The pipeline is delimited over three phases. Firstly, it is uploaded all the 1140 thermal matrices and images into Python. Then, the algorithm divides and matches the information for each patient with their respective diagnose (healthy or breast cancer). We used OpenCV python’s library for ROI extraction. Phase 1 supports the data pre-processing and augmentation for each of our proposed CAD systems. Phase 2 conveys the core of our scientific contribution, which is four sets of experiments further explained in the three following sections. This phase behaves depending on two auxiliary input functions: (i) a conventional training strategy and, (ii) a Bayesian optimization + conventional training. Lastly, Phase 3 evaluates the performance of our model using several types of metrics. Figure \ref{workflow} summarizes the pipeline of our methodology.

\subsection{Workflow description}

\begin{figure}[thp]
\begin{center}
\includegraphics[trim=0cm 54cm 23cm 0cm, clip, width=6in]{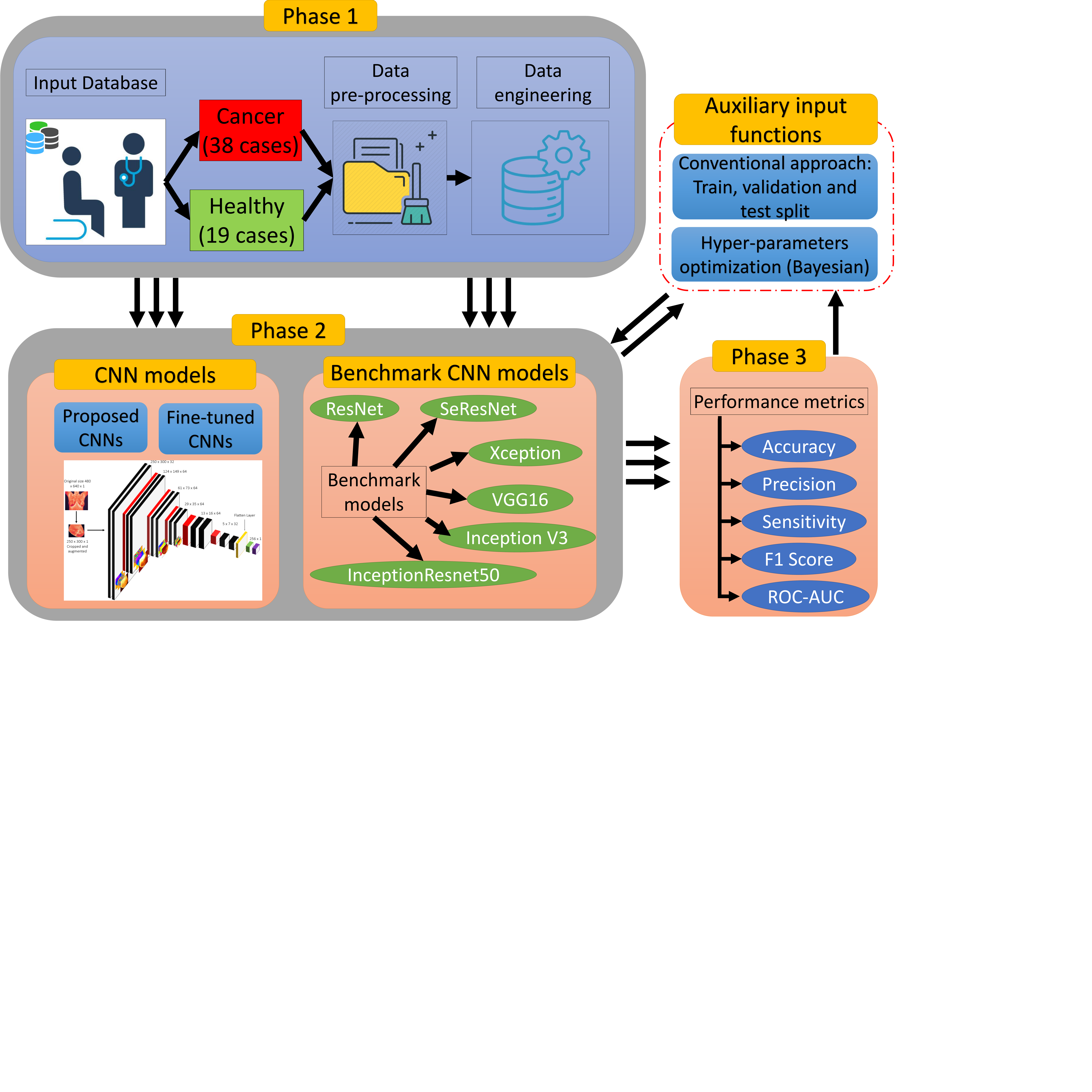}
\caption{The detailed workflow of fine-tuned and benchmark models. There are three phases: (1) describes the data acquisition, pre-processing and data augmentation; (2) shows the three core activities (baseline, benchmark and fine-tuned CNN models) and; (3) displays the performance metrics used for evaluating all the CNN architectures.}
\label{workflow}
\end{center}
\end{figure}

As mentioned before, our methodology is governed by Figure \ref{workflow} workflow. As the proposed methodology is interconnected, some experimental outputs become experimental inputs for other phases. Therefore, our experimental results are conducted by the following steps.  

\subsubsubsection{\textbf{Step 1: Database acquisition protocol}}

The DMR-IR database has a population of 57 patients, with an age between 21 and 80 years old; 19 patients are healthy and 37 present a malignant breast. The diagnostic has been prior confirmed via mammography, ultrasound and biopsies. The thermal images are captured with a FLIR thermal camera model SC620, which has a sensitivity of less than 0.04\degree C and captures standard -40\degree C to 500\degree C. Each infrared image has a dimension of 640x480 pixels; the software creates two types of files: (i) a heat-map file; (ii) a matrix with 640x480 points e.g. 307200 thermal points. Firstly, each patient undergoes thermal stress for decreasing the breast surface temperature and then twenty-image sequences are captured per 5 minutes. As a thermography test may be considerably affected when guidelines are not followed, the DMR-IR database followed the Ng \cite{46} and Satish \cite{12} acquisition protocol, which has been gathered jointly with physicians to ensure the database’s quality. Here, it is mentioned several standards that lead to high quality and unbiased thermal images. Firstly, each patient should avoid tea, coffee, large meals, alcohol and smoking before the test. Secondly, the camera needs to run at least 15 min prior to the evaluation, having a resolution of 100mK at 30\degree C; the camera at least should have 120x120 thermal points. Third, the recommended room’s temperature is between 18 and 25\degree C, humidity between 40\% and 75\%, carpeted floor, avoiding any source of heat such as, personal computers, devices that generate heat and strong lights. 

\subsubsubsection{\textbf{Step 2: Pre-processing and data augmentation}}

In the proposed algorithm at first, we have uploaded all the temperature matrices and mask to Python 3.7. After the data acquisition step, each breast ROI is segmented from the original gray-scale mask image, but depending on the patient’s health status, one (sick) or both (healthy) breasts are taken into consideration. In the pre-processing phase, we followed referenced methodologies \cite{61,64,63,68}, such as cropping, resizing and normalization of each thermal breast image. The product of this process is a thermal image with a size of 250x300 temperature points; consequently, we reduced by a quarter the computational cost. The data augmentation step conveys four types of image data generation: (i) horizontal and vertical flip; (ii) rotation between 0-45 degrees; (iii) 20\% zoom and; (iv) normalized noises, e.g. Gaussian. Algorithm \ref{MT_alg_1} presents a pseudo-code of the above-mentioned techniques, where the pre-processing and data augmentation methodologies are the same for all the performed experiments (excluding the fourth experiment). The fourth experiment measures the influence of data augmentation and database (DMR-IR) size on the CNN’s performance. It is important to mention that we assumed that the database’s acquisition protocol has been done rigorously \cite{12,46}, thus, minimizing the bias and obtaining a high-quality dataset. 

\begin{algorithm}
\caption{Data pre-processing \& data augmentation}
\label{MT_alg_1}
\begin{algorithmic}
\Procedure{Data Engineering}{}
\State $\textit{DBase} \gets \text{Input pre-processed database}$
\If {$\textit{Augmentation} = True$} select one or more:
\State $DBase \gets \text{DBase + horizontal or vertical flip}$
\State $DBase \gets \text{DBase + 0-45\degree ~image rotation}$
\State $DBase \gets \text{DBase + 20\% zoom}$
\State $DBase \gets \text{DBase + normalized noises}$
\State $\textit{YIELD (DBase)}$
\EndIf
\EndProcedure
\Procedure{Main Algorithm}{}

\BState \emph{main}:
\State $\texttt{(select one or more enhancement techniques)}$
\If {$\text{Scale Database}\to \textit{True}$}
\State $\textit{Dbase} \gets \text{Scale (}\textit{Dbase})$
\EndIf

\If {$\text{Crop Database}\to \textit{True}$}
\State $\textit{Dbase} \gets \text{Crop (}\textit{Dbase})$
\EndIf

\If {$\text{Resize Database}\to \textit{True}$}
\State $\textit{Dbase} \gets \text{Resize (}\textit{Dbase}) to 250x300$
\EndIf
\State $\textit{DBase augmented} \gets \text{YIELD (}\textit{Database pre-processed})$
\State \textbf{goto} \emph{Training Algorithm}.
\State \textbf{END}
\EndProcedure
\end{algorithmic}
\end{algorithm}

\subsubsubsection{\textbf{Step 3: Baseline and benchmark CNN models}}

\begin{figure}[thp]
\begin{center}
\includegraphics[trim=0cm 70cm 7cm 0cm, clip, width=6in]{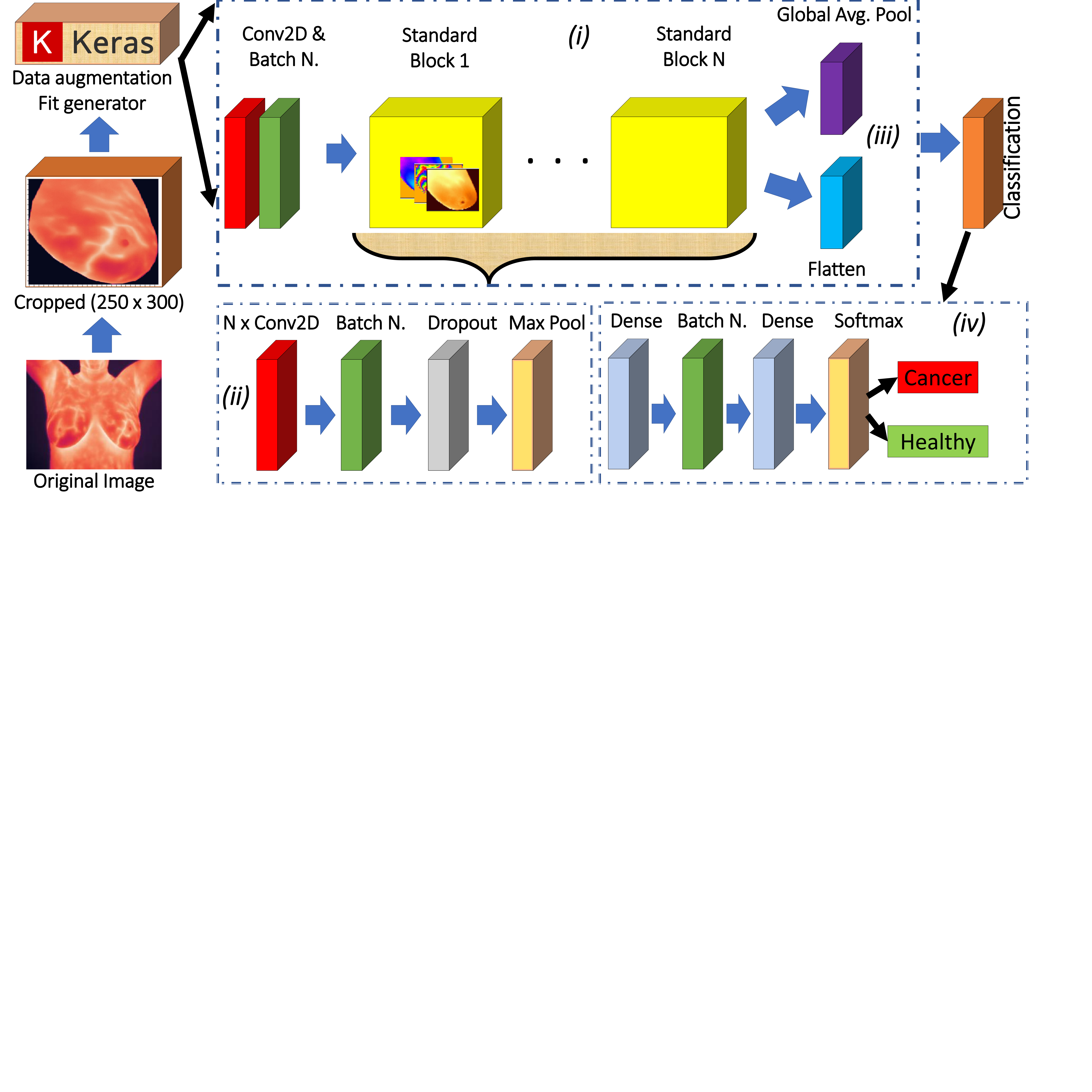}
\caption{Convolutional neural network architecture for the baseline and hyper-parameters optimization experiments. The (i) global overview of the model is composed of (ii) standard blocks (iii) coupled with two possible top layers (flatten or global average pooling). Then a classification block (iv) provides the breast cancer diagnostic.}
\label{CNN_block}
\end{center}
\end{figure}

During the second phase, the CAD system has two types of the auxiliary input function, as depicted in Figure \ref{workflow}. Firstly, we propose a database train, validation, and test split of 50/20/30, respectively. In order to match the methodologies done by other authors, we propose a CAD system that tests several baseline CNN architectures (proposed in Figure \ref{CNN_block}) under this training framework. In fact, some authors have obtained promising results using different methodologies and pre-processing techniques; nonetheless, other authors do not mention explicitly how is the database split \cite{2,51,55} during the training process; thus, there are doubts about the algorithms’ reliability and robustness when new cases will come. Contrary, we provide a detailed methodology starting from data preparation until the train/test phase, which guarantees the bias and overfitting minimization, even when new cases will come. Under that proposed training framework, we have tested several state-of-the-art CNN architectures: ResNet \cite{resnet}, SeResNet \cite{seresnet}, VGG16, Inception, InceptionResNetV2 \cite{inception_resnet}and Xception \cite{Xception}. Afterward, we made a comparison between the baseline and the state-of-the-art models, finding that simpler CNN models performed better than big CNN architectures; therefore, we applied optimization techniques to design an optimal CNN architecture that should perform better than experiment 1 and benchmark ones.

\subsubsubsection{\textbf{Step 4: CNN fine-tuning and hyper-parameters optimization}}

After comparing the baseline models with the two proposed approaches and the benchmark models, we decided to explore methodologies further to raise the performance of our CNN models; therefore, we propose a hyper-parameters Bayesian optimization based on a tree parzen estimator. Bayesian optimization is a probabilistic model-based approach for finding the global minimum of any function that returns a real-value metric (in our case F1-score). This methodology is also called a sequential model-based optimization because it builds a probabilistic model of an objective function that is based on past results. Each time the model receives new evidence, it updates the probability model (also called "surrogate model"), creating a new one with the last examples. The longer the algorithm runs, the closer the surrogate function comes to resembling the actual objective function. We implemented on Python 3.7 the tree parzen estimator (TPE) using the HyperOPT library \cite{hyperopt}. In fact, there are four key phases in order to build a TPE pipeline:

\begin{itemize}
    \item Firstly, it is necessary to define a domain space that will change depending on the model’s evolution. The domain space may vary depending on the past results or the type of optimizer,
    \item The optimization algorithm. In this case, a TPE bayesian model,
    \item The objective function receives a set of hyper-parameters then, create a machine-learning model (here, a CNN), 
    \item An evaluation metrics function receives a set of predicted values and real labels (cancer or healthy). Next, it returns metrics like accuracy, precision, sensitivity, F1-score and ROC-AUC.
\end{itemize}

The CAD system's performance tells how close is the system to correctly diagnose whether a patient is having the disease and the ones who do not. In our experimentation, the people carrying a malignant breast are the true class and the ones healthy are the false class. Therefore, the easiest way to summarize a CAD system’s performance is with evaluation metrics. We demonstrate the performance of our CNN models with several metrics such as accuracy, precision, sensitivity, F1-score and ROC-AUC. However, we have chosen F1-score rather than accuracy. On the one hand, the F1-score takes both false positives and false negatives into account; on the other hand, accuracy takes true positives and true negatives. The false-negative (FN) is a result that indicates a person does not have breast cancer when the person actually does have it. The false positive (FP) is a result that indicates a person does have breast cancer when the person actually does not have it. The below equations depict the proposed metrics. 

\noindent\begin{tabularx}{\textwidth}{XX}
  \begin{equation}
  precision=\frac{TP}{TP+FP}
    \label{equation_precision}
  \end{equation} &
  \begin{equation}
  Sensitivity=\frac{TP}{TP+FN}
    \label{equation_recall}
  \end{equation}
\end{tabularx}
\begin{equation}
F1-score=2 . \frac{Precision * Sensitivity}{Precision+Sensitivity}
\label{equation_f1_score}
\end{equation}

\textbf{Remark 1:} Thence, knowing that the early diagnosis of breast cancer is crucial for the patient’s survival, the FN and FP are much more crucial parameters for a CAD system, in order to diagnose the disease and reduce the mortality index. Additionally, F1-score deals with the imbalanced class distribution problem where accuracy does not. Thus, in the biomedical area and specifically in breast cancer diagnosis, it is much more convenient than the F1-score. This metric is also recognized as the harmonic mean between precision (Equation \ref{equation_precision}) and recall (Equation \ref{equation_recall}) as depicted in Equation \ref{equation_f1_score}. Finally, in the case of an equal F1-score in two CNN models, we have chosen the one with greater sensitivity, as it takes into account the FN.

\section{Experimental setup up and results}
\label{sec:results}

This section delimits our study but also conveys the main finding of our studies and the top CNN models obtained from empirically experimentation and hyper-parameters optimization. Additionally, we study the influence of data augmentation and database size on the CNNs performance. 

\subsection{Experimental setup}

The experimental set up is composed of four consecutive experiments, explained as follows: firstly, we have developed a baseline CNN models following Figure \ref{CNN_block} architecture. Our algorithm was trained with a Tesla K80 GPU unit, free and available in Google Colab. The training framework splits the database as follows, 50\% train, 20\% validation and 30\% test sets, following the Approach methodology from Figure \ref{workflow}. All the experiments had batch normalization layers, ReLU activation function and we tested several optimizers such as Adam, RMSprop and SGD. Throughout the experiments, it has been tested several architectures varying the number of Conv2D layers, the dropout rate and the number of units in the last dense connected layers. The input image size is 250x300 temperature “pixels”. The training process has been done under the mini-augmented training batches (32 augmented images per step), with 50 steps per epoch (50 evaluation of 32 instances, per epoch), and 40 epochs in total. Finally, we summarized the results and we selected the best model based on performance metrics and execution time. As a second part of the first set of experiments, we have done a crucial change in the database splitting from here and now on (Approach 2 from Figure \ref{methodology_1}). Instead of splitting the whole database immediately, we have done a balanced splitting by patient thus, 39 patients for the train (780 images) and 17 for the test set (340 images). Again, we summarized the results and we selected the best model. 

\begin{figure}[thp]
\begin{center}
\includegraphics[trim=0cm 78cm 45cm 0cm, clip, width=6in]{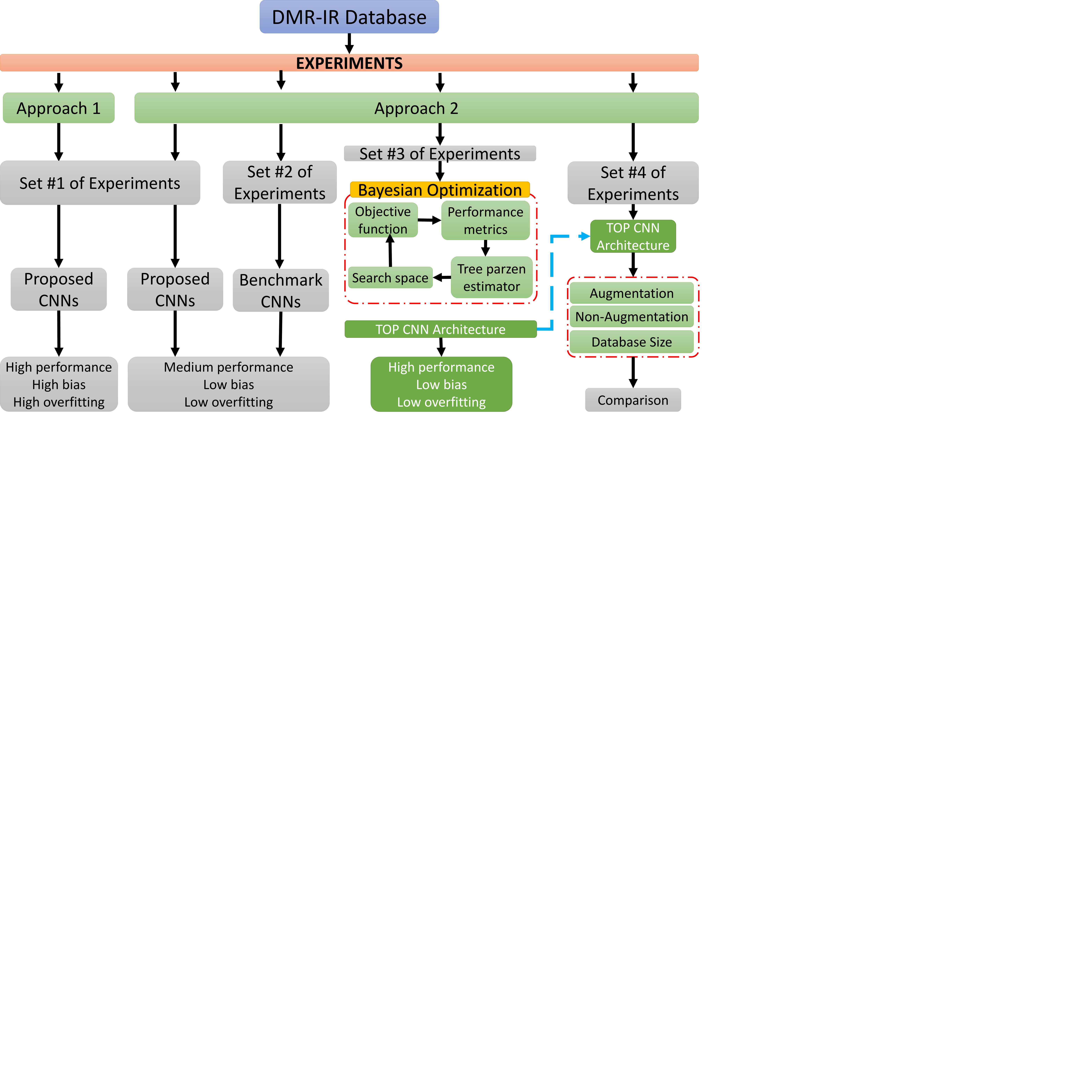}
\caption{Experiment proposed workflows. Experiment 1 compares the CNN performance with Approach 1 and Approach 2 from Figure \ref{methodology_1}. Experiment 2 presents a CNN state-of-the-art benchmark. Experiment 3 applies a Bayesian optimization to determine the optimal CNN architecture. Experiment 4 defines the influence of data pre-processing and data augmentation on the DMR-IR database.}
\label{methodology_2}
\end{center}
\end{figure}

The significant change in the performance metrics found during the first set of experiments motivated us to search alternative CNNs architectures to further improve our proposed CAD system. Consequently, the second set of experiments compares state-of-the-art CNN architectures with our previous results from the first experimentation. We prearranged several novel and up-to-date CNNs architectures. We keep each original CNN architecture, but we changed the top layer by a Flatten or global average pooling (GAP) layer, followed by two dense layers of 1024 units and a two-unit dense layer with Softmax activation function. It is important to recall that all the performance metrics obtained from here and now on are based on blind test samples, i.e. samples that have not been seen by the models during the training. Similarly, we have applied three callback functions in Keras, those are: (i) model checkpoint to save the weights of our top model; (ii) learning rate scheduler to apply a decay learning rate after each epoch and; (iii) early stopping monitor to reduce the overfitting and stop the training process when the model has stopped to learn. The second set of experiments brought a main conclusion: the simpler the CNN model, the higher the performance metrics. Thus, as discussed in Section \ref{sec:methodology}, we decided to apply optimization techniques to find an optimal CNN architecture.  

The third set of experiments aims to find the optimal CNN architecture; particularly we have developed a Bayesian optimization. As explained in Section \ref{sec:methodology}, the optimization algorithm needs a search space where to examine and chose the optimal hyper-parameters. Consequently, we defined as hyper-parameters (based on Figure \ref{CNN_block}): (i) minimum and maximum number of blocks; (ii) number of Conv2D layers per each block; (iii) number of filters per Conv2D layer; (iv); (v) type of optimizer; (vi) kernel size; (vii) pooling layer size; (viii) batch normalization; (ix) dropout rate; (x) number of dense units in the last two layers; (xi) top layer type. In the final step, the optimization algorithm chooses from fifty CNN models, the one with highest F1-score. 

The fourth set of experiments takes as input the top model from the Bayesian optimization phase and applies different training frameworks. Our main goal is to measure the CNN’s performance under different training scenarios. On the one hand, we compared the results with and without data augmentation; on the other hand, we varied the database split ratio between train, validation and test datasets. As a result of this experimentation, we are able to advise an optimized population size needed to replicate each result obtained throughout this study. The training framework is as follows, we (1) selected randomly ten patients, five healthy and five with the disease; (2) trained five CNN models with 10, 20, 30, 40 and 47 patients with an 80/20 train and validation sets split; (3) evaluated the performance of each of the five models. Afterward, we repeated the process from (1) to (3) four times, such as a “k-fold” cross-validation and we obtained the mean of each performance metric. A general overview of these proposed set of experiments could be found in Figure \ref{methodology_2}. 

\subsection{Experimental results}

The purpose of the first set of experiments was to establish baseline CNN models, which tells the advantages of MLT over texture and statistical features in breast thermography. As mentioned in Section \ref{sec:methodology}, a CNN has various hyper-parameters that influence the learning during the training framework, resulting in satisfactory or unacceptable results; thus, it is imperative to find the best combination of these parameters to ensure the CAD system reliability and robustness. In the first set of experiments, the leading parameters were: (i) number of CNN layers and filters; (ii) batch normalization and dropout rate; (iii) optimizer. Despite the training methodology is not a cataloged as hyper-parameter, it was important to separate the results depending on it. As discussed in the experimental setup, during the first set of experiments, we tested two different database split approaches (see Figure \ref{methodology_1}) in four CNN architectures changing the set of hyper-parameters randomly. In total, the algorithm runs a forty epochs simulation per model, which summed 200 epochs per database split methodology. Table \ref{tab:exp_1} summarizes all the CNN performance metrics with each proposed splitting approach, where CNN$i$, but $i=1,2,3,4$. We implemented a dropout rate to increase the model’s robustness by dropping out inputs from one layer to the next one. Additionally, it is set a ten-epoch early stopping callback to reduce the overfitting during the training process. 

\begin{table}[t]
\centering
\caption{Comparison of five performance metrics on four CNN architectures. The hyper-parameters were given empirically and it has been tested Approach 1 (biased) and Approach 2 (unbiased) database split methodology from Figure \ref{methodology_1} for each CNN.}
\label{tab:exp_1}
\begin{tabular}{|p{1.2cm}p{1cm}cp{1cm}p{01cm}p{0.5cm}p{0.5cm}p{0.5cm}p{0.5cm}p{0.5cm}p{1.5cm}|}
\toprule
\parbox[t]{1.2cm}{Model} & \parbox[t]{1cm}{Class} & \parbox[t]{2.5cm}{\centering Architecture (Num. of blocks, num of layers)} & \rotatebox[origin=c]{90}{Optimizer} & \parbox[t]{1cm}{Top Layer} & \rotatebox[origin=c]{90}{Accuracy} & \rotatebox[origin=c]{90}{F1 score} &  \rotatebox[origin=c]{90}{Precision} & \rotatebox[origin=c]{90}{Sensitivity} & \rotatebox[origin=c]{90}{ROC-AUC} & \parbox[t]{1.5cm}{\raggedright Time per epoch (s)}\\
\midrule
\textbf{CNN 1} & Biased & (5,3) & SGD & Flatten & \textbf{0.99} & \textbf{0.99} & \textbf{0.99} & \textbf{0.98} & \textbf{0.99} & \textbf{26} \\
 & Unbiased & (5,3) & SGD & GAP & \textbf{0.86}& \textbf{0.87} & \textbf{0.84} & \textbf{0.90} & \textbf{0.85} & \textbf{30} \\
\cline{2-11}
&&&&&&&&&&\\
CNN 2 & biased & (6,4) & SGD & Flatten & 0.99 & 0.99 & 0.99 & 0.97 & 0.99 & 26 \\
 & unbiased & (6,4) & Adam & GAP & 0.83 & 0.82 & 0.92 & 0.75 & 0.84 & 29 \\
\cline{2-11}
&&&&&&&&&&\\
CNN 3 & biased & (7,4) & Adam & GAP & 0.92 & 0.92 & 0.94 & 0.90 & 0.92 & 23 \\
& unbiased & (7,4) & Adam & GAP & 0.85 & 0.86 & 0.83 & 0.89 & 0.84 & 21 \\
\cline{2-11}
&&&&&&&&&&\\
CNN 4 & biased & (4,3) & Adam & Flatten & 0.89 & 0.89 & 0.92 & 0.87 & 0.90 & 21 \\
 & unbiased & (4,3) & Adam & GAP & 0.86 & 0.87 & 0.90 & 0.84 & 0.87 & 25\\
\bottomrule
\end{tabular}
\end{table}

The model CNN 1 yielded the best performance in both cases, approaches 1 and 2. It is important to recall that we selected the best model based on \textbf{Remark 1} from Section \ref{sec:methodology}. In the first instance, CNN1 yielded 99\% accuracy, 99\% precision, 98\% sensitivity, 99\% F1 Score and 99\% ROU-AUC. Nevertheless, the second instance showed a lower performance with an 88\% accuracy, 88\% precision, 91\% sensitivity, 89\% F1-score and 88\% ROU-AUC. Indeed, each CNN model has better result when using Approach 1, because there is a high probability that the CNN models under this database split methodology had images from the same patient in both datasets, train and test. In other words, images from the twenty-image sequences pertaining to a given patient could be belonging to both, the train and test set (or validation set) simultaneously. The baseline results obtained from the first set of experiments suggested that more experimentation was needed in order to reach a CAD system with high performance, low bias and low overfitting. Therefore, the idea of searching for better CNN architectures concluded in a new set of experiments based on state-of-the-art CNN architectures, which might be capable of overcoming the weaknesses encountered during the first set of experiments. 

The second set of experiments involves the benchmark of state-of-the-art CNN architectures such as ResNet, SeResNet, Inception version 3, VGG16, InceptionResNet V2 and Xception. Table \ref{tab:exp_2} exhibits the performance metrics for all the proposed models. Generally, these cutting-edge CNN models are well optimized in architecture, but come at a cost of high number of parameters; indeed, higher than the models from experiment 1. We kept the database split methodology (Approach 2), datasets proportion, the number of training epochs and the early stopping callback all along the second set of experiments. 

\begin{table}[htb]
\centering
\caption{Summary of performance metrics of each CNN model from the second set of experiments (benchmark) and the top model from the first set of experiment (Approach 1).}
\label{tab:exp_2}
\begin{tabular}{|ccccccc|}
\toprule
Model & Accuracy & F1-score & Precision & Sensitivity & ROC AUC & Time per epoch (s) \\
\midrule
\textbf{SeResNet18} & \textbf{0.90} & \textbf{0.91} & \textbf{0.91} & \textbf{0.90} & \textbf{0.90} & \textbf{30} \\
SeResNet34 & 0.86 & 0.86 & 0.91 & 0.81 & 0.86 & 35 \\
SeResNet50 & 0.82 & 0.81 & 0.85 & 0.78 & 0.82 & 42 \\
ResNet50 & 0.79 & 0.77 & 0.90 & 0.68 & 0.80 & 30 \\
VGG16 & 0.90 & 0.89 & 0.85 & 0.94 & 0.90 & 22 \\
InceptionV3 & 0.80 & 0.80 & 0.82 & 0.78 & 0.80 & 21 \\
InceptionResNetV2 & 0.65 & 0.72 & 0.93 & 0.59 & 0.72 & 44 \\
Xception & 0.90 & 0.89 & 0.89 & 0.90 & 0.90 & 30 \\
\textbf{Top CNN (App 2.)} & \textbf{0.86}& \textbf{0.87} & \textbf{0.84} & \textbf{0.90} & \textbf{0.85} & \textbf{30} \\
\bottomrule
\end{tabular}
\end{table}

Table \ref{tab:exp_2} shows the performance metrics for each individual model during the forty-epoch training. We tested for each CNN model both top layers, flatten and GAP layers. In all the cases, the GAP predominated with higher performance e.g. the Inception V3 CNN model had a 30\% improvement of F1-score when using GAP (not shown in Table \ref{tab:exp_2}) rather than flatten layer. During the experimentation, GAP layers and Adam (rather than RMSProp or SGD) optimizer yielded much better results; thus, we implemented this set of hyper-parameters for all the proposed state-of-the-art CNN models. The preeminent CNN model was the SeResNet18 with a 90\% accuracy, 91\% precision, 90\% sensitivity, 91\% F1-score and 90\% ROU-AUC. We reach a breakpoint during this experiment, which suggests that simpler CNN models are better for the DMR-IR database. 

\begin{figure}[thp]
\begin{center}
\includegraphics[trim=0cm 90cm 55cm 0cm, clip, width=5.2in]{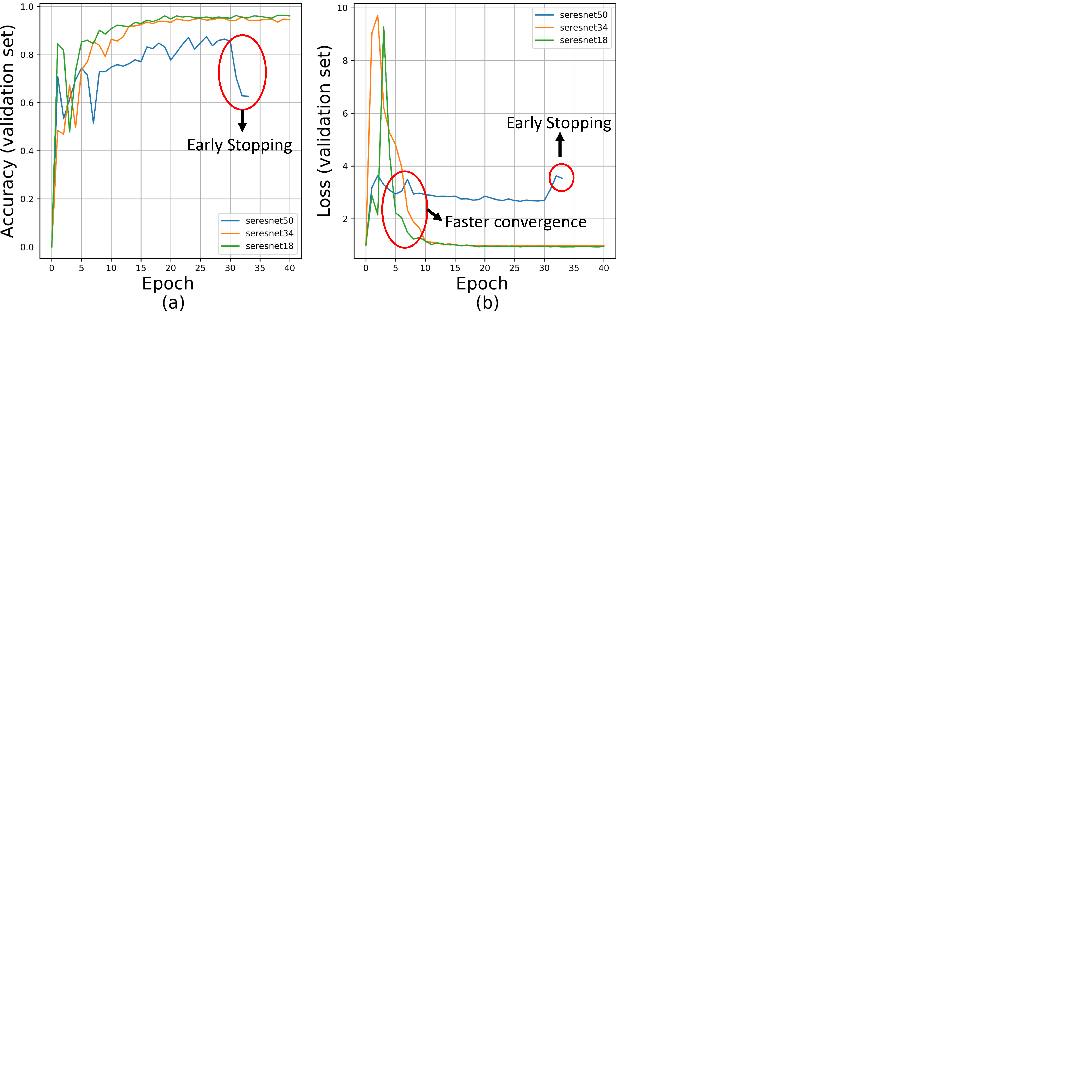}
\caption{Validation datasets performance of SeResNet18, SeResNet34 and SeResNet50 during training for 40 epochs. Model’s (a) accuracy and (b) losses in the validation dataset.}
\label{serenest_exp}
\end{center}
\end{figure}

As the complex the CNN architecture, the worse the model’s performance, the Figure \ref{serenest_exp} plots the accuracy and loss results over the training process of three SeResNet CNN models (presented in Table \ref{tab:exp_2}). Each plot represents one CNN architecture with a GAP layer followed by two fully connected layers of 1024 units each. Likewise, some CNN did not reach the forty-epoch goal due to the early stopping callback, which allows us to stop training when the model has stopped to learn. We applied L2 regularization after detecting overfitting in some CNN models. From a general point of view, most of the state-of-the-art CNN architectures were not as regular as the ones presented in experiment 1 (suggested CNN architecture by the authors); we believe that these models are better for small datasets such as DMR-IR database, which it is a binary classification problem (healthy or malignant breast). Contrary, Table \ref{tab:exp_2} CNN benchmark models are for multi-class classification on huge databases like ImageNet. In conclusion, these benchmark results motivated us to pursuit optimization techniques to obtain the best CNN architecture.  

A hyper-parameter is a parameter that controls the learning process of a given algorithm. The acquisition process of biomedical databases in most of the cases is expensive and should follow strict guidelines in order to obtain high-quality data; thus, these databases usually are both, small and unbalanced. We implemented a Bayesian optimization (explained in Section \ref{sec:methodology}) of CNN hyper-parameters to deal with these problems. Firstly, our algorithm chooses a set of hyper-parameters from the proposed search space of Table \ref{tab:search_space}. Secondly, the objective function creates a CNN model based on these learning and architectural parameters. The TPE algorithm provides a result based on the current and past performance metrics of the previous models.

\begin{table}[htb]
\centering
\caption{Search space for Bayesian optimization of CNN hyper-parameters with a tree parzen estimator. Figure \ref{CNN_block} delimits the CNN architecture.}
\label{tab:search_space}
\begin{tabular}{|c|ccc|}
\toprule
& Hyper-parameter & Min & Max \\
\midrule
\multirow{7}{*}{\rotatebox[origin=c]{90}{\parbox[c]{1cm}{\centering Quantitative}}} & Number of blocks & 2 & 4 \\
 & 2D Conv. layers per block & 2 & 5 \\
 & Number of filters per layer & 64 & 512 \\
 & Kernel size (n x n) & 2 & 4 \\
 & Pooling layer size (n x n) & 2 & 3 \\
 & Dense Layers (Num. units) & 256 & 1024 \\
 & L2 regularizer & 0 & 0.2 \\
 \midrule
 \multirow{5}{*}{\rotatebox[origin=c]{90}{\parbox[c]{1.5cm}{\centering Qualitative}}} & Optimizer type & \multicolumn{2}{c|}{Adam, SGD, RMSProp} \\
  & Droupout & \multicolumn{2}{c|}{Yes, No} \\
  & Batch Normalization & \multicolumn{2}{c|}{Yes, No} \\
 & Type of activation function & \multicolumn{2}{c|}{Elu, ReLU} \\
 & Type of top layer & \multicolumn{2}{c|}{Flatten or GAP} \\
\bottomrule
\end{tabular}
\end{table} 

The best result was obtained when we associated GAP or flatten layers with a dropout rate between 0 and 0.3, but also with 6 or 7 CNN blocks. To demonstrate the success of the hyper-parameters optimization, our CNN top model yielded a 92\% accuracy, 98\% precision, 87\% sensitivity, 92\% F1-score and 92\% ROU-AUC as classification metrics in the DMR-IR database. Non-conventional techniques such as our optimization algorithm are able to increase the CAD system’s performance; specifically, the mean accuracy score raised by a 6\% and 8\% compared with experiments 1 and 2, respectively. The hyper-parameters optimization problem was targeted as a minimization problem; despite an overall of 207.360 possible combinations of hyper-parameters, we tested fifty different sets. Table \ref{tab:exp_3} presents a summary of the top models obtained during the Bayesian optimization and during experiments 1 and 2. We proposed three CNN models, where CNN-Hyp$i$, but $i=1,2,3$. 

\begin{table}[htb]
\centering
\caption{Performance metrics comparison between the top CNN models from the third set of experiments (Bayesian optimization) and experiment 1 and 2. The database splitting follows the Approach 2 from Figure Figure \ref{methodology_1}.}
\label{tab:exp_3}
\begin{tabular}{|cccccccccc|}
\toprule
\parbox[t]{1.5cm}{Model} & \parbox[t]{2.5cm}{\centering Architecture (Num. of blocks, num of layers)} & \rotatebox[origin=c]{90}{Optimizer} & \parbox[t]{1cm}{Top Layer} & \rotatebox[origin=c]{90}{Accuracy} & \rotatebox[origin=c]{90}{F1 score} &  \rotatebox[origin=c]{90}{Precision} & \rotatebox[origin=c]{90}{Sensitivity} & \rotatebox[origin=c]{90}{ROC-AUC} & \parbox[t]{1.5cm}{\raggedright Time per epoch}\\
\midrule
\textbf{CNN-Hyp 1} & (6,3) & RMSProp & Flatten & \textbf{0.94} & \textbf{0.91} & \textbf{0.92} & \textbf{0.92} & \textbf{0.92} & \textbf{40} \\
CNN-Hyp 2 & (6,4) & SGD & GAP & 0.99 & 0.83 & 0.90 & 0.90 & 0.91 & 64 \\
CNN-Hyp 3 & (7,2) & Adam & flatten & 0.98 & 0.87 & 0.92 & 0.92 & 0.92 & 24 \\
\cline{2-10}
SeResNet18 & - & Adam & 30 & 0.91 & 0.90 & 0.91 & 0.90 & 0.90 & 30 \\
Top CNN (App 2.) & (5,3) & SGD & GAP & 0.84 & 0.90 & 0.87 & 0.86 & 0.85 & 30\\
\bottomrule
\end{tabular}
\end{table}

\begin{figure}[thp]
\begin{center}
\includegraphics[trim=1.5cm 1.1cm 3cm 2cm, clip, width=6in]{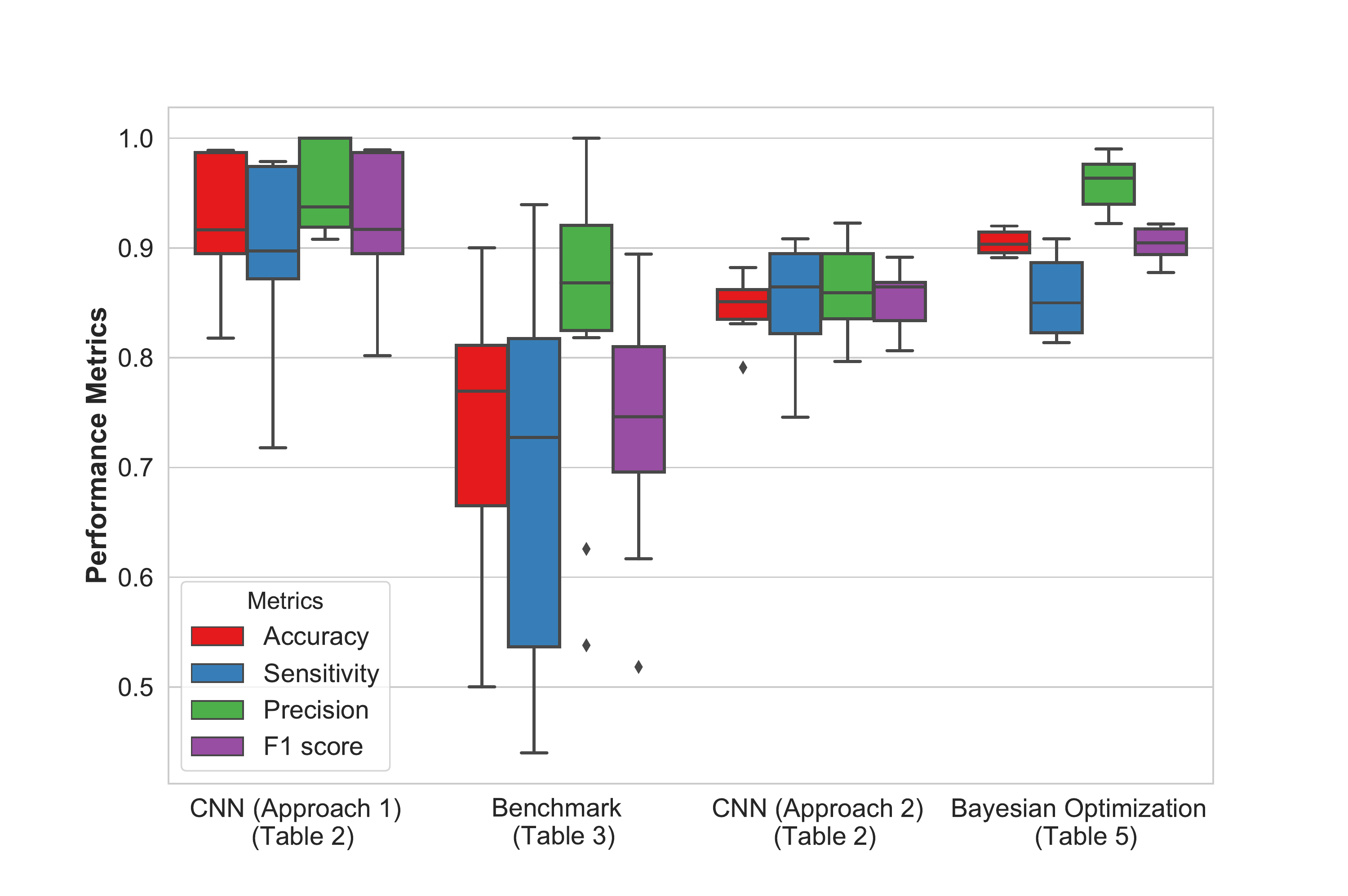}
\caption{Summarized box-plot of performance metrics for all sets of experiments. Experiment 1 uses the database split approach 1 and 2 from Figure \ref{methodology_1}, respectively. Experiment 2 corresponds to CNN benchmarking models. Experiment 3 shows the top results obtained throughout the Bayesian optimized CNN models.}
\label{comparison_exp}
\end{center}
\end{figure}

To get a general overview of the results of our proposed experiments, Figure \ref{comparison_exp} summarizes the averaged performance metrics from experiments 1 to 3. We decided to show in this figure the experiments in ascending order: firstly, despite experiment 1 successful performance, we concluded that it was biased and over-fitted due to the training framework (database split Approach 1), weakening the models’ robustness. Secondly, the benchmark experimentation had higher dispersion in comparison with the other set of experiments, diminishing the model’s reliability. Finally, we measured the evolution in the performance metrics from the empirically given hyper-parameters (CNN with Approach 2) and the optimized set of hyper-parameters (Bayesian Optimization) for obtaining the topmost CNN architecture. It is important to note that the Bayesian optimization experiment displays an average increase of 7\% in F1-score compared with experiment 1 (App. 2) and the benchmark experiments. 

\subsection{Influence of data augmentation and database size }

The reliability and availability of databases for breast cancer diagnosis using thermography are major challenges nowadays. Consequently, this section brings guidance for new researchers in breast thermography, dealing with databases’ size issues and the role of data augmentation. In the fourth and final set of experiments, we measured the influence of data augmentation techniques and database size in the models’ performance for the DMR-IR database. In addition, each performed experiment has been tested with and without data augmentation techniques.

\begin{figure}[thp]
\begin{center}
\includegraphics[trim=5cm 1.5cm 5cm 3cm, clip, width=6in]{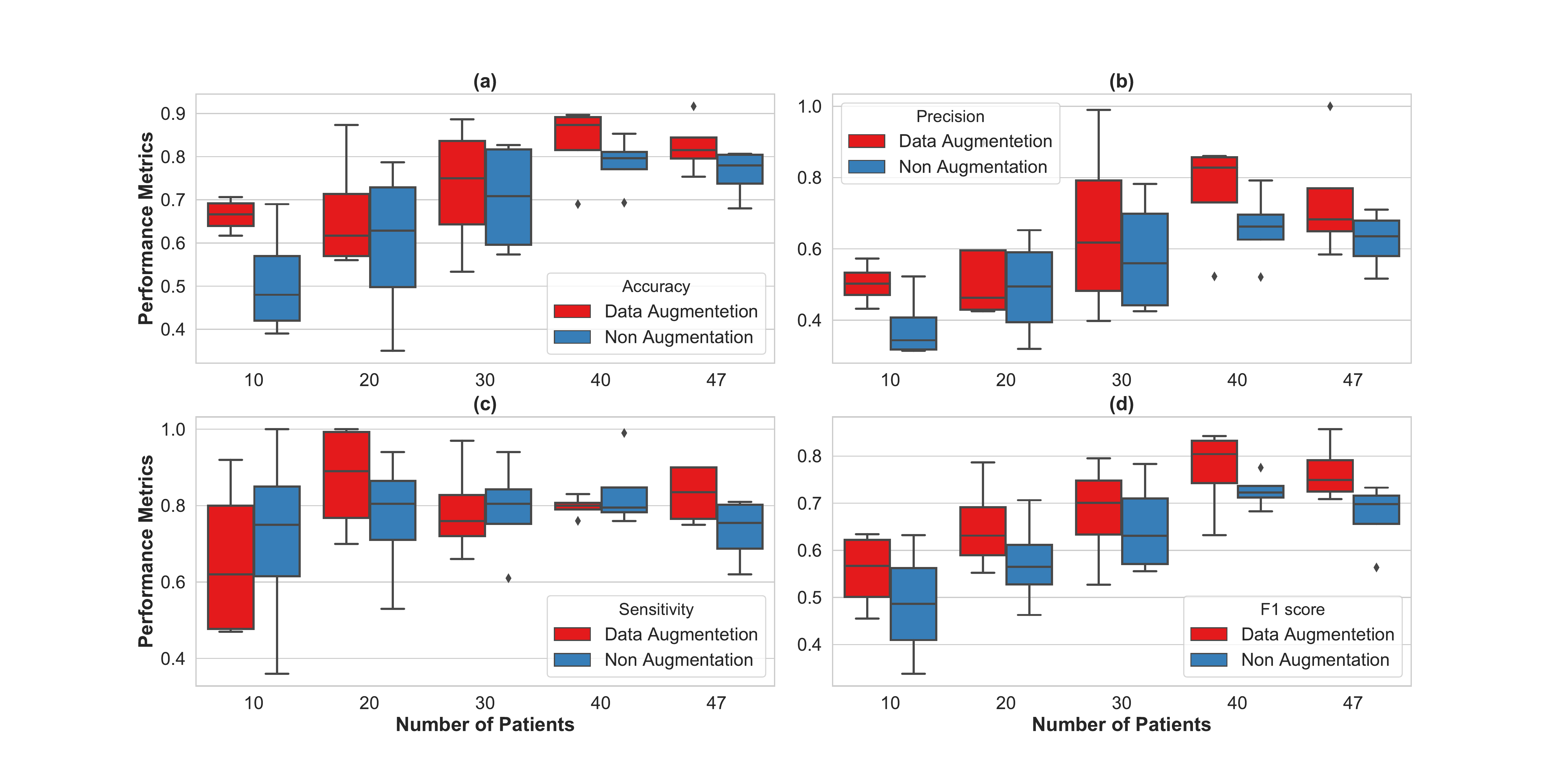}
\caption{CNN averaged performance metrics over number of patients taken from the DMR-IR database, with and without data augmentation. The test set has randomly chosen ten patients for all cases (4 folds). (a) Accuracy, (b) Precision, (c) Sensitivity and (d) F1-score.}
\label{data_aug_comparison}
\end{center}
\end{figure}

Figure \ref{data_aug_comparison} plots the averaged performance metrics varying the train/validation dataset size from 10 to 47 patients, with and without data augmentation techniques. We decided to choose an averaged performance (4 fold of metrics) rather than one set metrics because, the averaged performance decreases the excessively high bias and variance of CNNs in unseen data. We followed the k-fold cross-validation methodology, but instead changing the train/validation set, we tested four different “test sets” i.e. four test folds. The main objective during this set of experiments is to prove the advantage of data augmentation rather than no data augmentation. Section \ref{sec:discussion} discuss the main insights about the results presented throughout Section \ref{sec:results} and some recommendations towards future works.

\section{Discussion and conclusion}
\label{sec:discussion}

The results presented throughout Tables \ref{tab:exp_1} to \ref{tab:exp_3} provide a general overview of our contribution in (i) comparing the CNNs’ performance over different database split methodologies in the DMR-IR database; (ii) providing a new methodology that highly decrease the overfitting and biasing during the training process of CNNs for this database; (iii) a benchmark comparison of state-of-the-art CNN models for the DMR-IR database. In addition, we (iv) demonstrated the benefits of hyper-parameters optimization for fine-tuning CNN architecture and; (v) measured the influence of data augmentation techniques and datasets sizes in the DMR-IR database. Hence, it is becoming as baseline information for future works in either breast thermography databases or conception of new ones. 

During the last years, there have been a demand for high quality, cheap, and reliable CAD systems for a breast cancer diagnosis; but specifically, early detection. CNN-based CAD systems for thermography stands as one methodology that could satisfy those requirements. However, the lack of public databases has limited the studies towards thermography. In fact, the only public and free database is the DMR-IR \cite{52,53}. We assumed that DMR-IR is one of the main databases in thermography due to both, his high quality (fulfilling the standard acquisition protocols \cite{12, 46}) and his acceptance in the research community (see Table \ref{tab:summary_studies}). Nonetheless, when referring to past studies, it has been becoming almost impossible to compare the results impartially from study to study due to difference in the training framework, database size, datasets split ratio (between train/validation/test sets), normalization techniques and types of CAD system (texture and statistical features or CNNs).

Despite past studies that have diverged in the database sizes, we have seen two experimental methodologies. On the one hand, some authors have used texture and silhouette features coupled with machine learning techniques \cite{51,53,55,62,63,64,67} to detect whether a patient does have cancer. On the other hand, a pair of studies use machine-learning techniques straightforward with the DMR-IR thermal images \cite{65,66}. As far as it is known, gathering texture, silhouette and statistical features demand much more time than applying MLT e.g. CNN. In addition, algorithms based on these features need more time and resources to reach the required reliability and robustness, this due to the large intra-class appearance variations triggered by changes in illumination, rotation, scale, blur, noise, occlusion, etc. Likewise, the main idea of CNN-based CAD system is to minimize the rate of pre-processing and data management needed prior to conceiving a robust machine-learning system, focusing further on the CNN architecture itself. In other words, the developing time of a fully operational CAD system based on CNNs is fewer compared to one based on texture and statistical features. 

As this study is the employing of parallel MLT rather than algorithms based on texture and statistical features from thermal images, we have chosen CNN due to their performance in spatial-databases. High-level APIs such as Keras from Tensorflow \cite{tensorflow} allows the rapid development of robust CNN architectures. We focused our first set of experiments on demonstrating the impact and consequences of the database splitting methodology over the training. On the one hand, the first set of Table \ref{tab:exp_1} CNN models follows the Approach 1 for database split methodology, where some authors have presented results using a small \cite{55,61}, medium \cite{63} and full \cite{64,65} part of the DMR-IR database. The main concern with this methodology is the high performance achieved during training e.g. our top model has an accuracy, F1-score and precision of more than 98\%. On the other hand, the second set of Table \ref{tab:exp_1} CNN models use a more robust training framework, which all images/sequences are pertaining to a given patient either, all belong to the training or the testing set (or validation set); thence, minimizing the bias and over-fitting. Although these models yielded an average accuracy and F1-score of 84\% and 85\%, a thermography-based CAD system requires higher performance to overcome techniques like mammography. Generally, in the first set of experiments, CNN models with flatten layer and SGD optimizer had better results when training under Approach 1; contrary, mixing GAP layer and Adam optimizer yielded higher performance under Approach 2. No one before Fernández-Ovies et al. \cite{66} has made a benchmark of state-of-the-art CNN architectures such as ResNet or VGG, employing the DMR-IR database. Likewise, as a second general contribution, Table \ref{tab:exp_2} and Figure \ref{serenest_exp} depict a benchmark study of several state-of-the-art CNN architectures. In previous studies \cite{65,66}, the essential contribution was not a CNN benchmark study but rather the employment CNNs as core MLT for their CAD systems. We noticed that CNNs models with Inception modules (e.g. Inception V3 and InceptionResnet) had a lower performance because these architectures have many weights and parameters to tune, so we arrive to a breakthrough conclusion: the patterns in the DMR-IR thermal images are not too complex to be generalized by a CNN. In consequence, the complex the CNN (width, depth and number of filters), the hard to generalize the DMR-IR thermal images. In order to verify these conclusions, we developed specific experimentation using several SeResNet \cite{seresnet} but changing the number of residual layers. In the first case, we obtained an 81\% accuracy, 85\% precision, 78\% sensitivity and 81\% F1-score with a SeResNet50, but following our assumption that the simpler the model, the better; we tested a SeResNet34 and SeResNet18. Consequently, we obtained a 9\% accuracy, sensitivity and F1-score improvement when using the simpler model –SeResNet18-. To further prove our hypothesis, Figure \ref{serenest_exp} shows the validation accuracy and losses versus epoch during the training period, where the simpler the model, the faster the model converged. 

The idea of simpler a CNN model yields better results that motivated to seek non-conventional techniques to improve our CNN models. Specifically, we implemented a Bayesian Optimization based on a TPE to obtain the optimal CNN architecture (see Figure \ref{CNN_block}) from the search space suggested in Table \ref{tab:search_space}. As mentioned before, the top models obtained throughout the optimization performed much better than the experiment 1 and 2. In general, the flatten layer achieved better results than GAP, the SGD needed more processing time than RMSprop and Adam optimizer, but in all the cases, the results were comparable. To summarize, we plotted in Figure \ref{comparison_exp} the averaged results per experiment and per metric, from experiments 1 to 3. We deduced from Figure \ref{comparison_exp} that experiment 1 App. 1 obtained the best performance metrics but at the cost of high bias and over-fitting during the training; contrary, the App. 2 yielded high performance, but the CNN architecture was given empirically. The average of experiment 2 produced a high variance in the box-plots, because some benchmark CNN models achieved high performance, but other who does not. Finally, experiment 3 collects all the positive things such as low variance, low bias and low overfitting on the averaged performance metrics on three CNN models; moreover, rather than gives an empirical architectures to this models, we opted to apply a Bayesian optimization that yielded the optimal architecture, which overcomes all the previous CNN models. 

Despite the main advantages of CNNs, one of the main known drawbacks in MLT-based CAD systems is the quantity of available data, specifically in our case, breast thermal images. In most of the circumstances, gather more data demands expensive and rigorous protocols, which should ensure the databases’ high quality and reliability. Consequently, we targeted this problem inversely following the performance evolution of several CNN models when was applied data augmentation and when was altered the database size i.e. the number of patients. The Figure \ref{data_aug_comparison} summarizes the accuracy, sensitivity, precision and F1-score of the proposed before-mentioned comparison and also explained in Section \ref{sec:results}. Therefore, the fourth and last set of experiments suggested –as expected- that the larger the database size (i.e. from 10 to 47 patients), the more the CNNs generalize the data and the more the performance increase. When the performance increased, the CNN models were more regular, having therefore less variance, as can be seen in Figure \ref{data_aug_comparison} (d). Overall, the data augmentation techniques during all the simulations performed much better than no data augmentation; for instance, the mean F1-score in all cases was at least 10\% higher. If we compare the F1-score (Figure \ref{data_aug_comparison} (d)) of the experiments with databases sizes of 10, 20 and 30 patients, we conclude that a CNN model which uses data augmentation techniques requires 50\% less number of patients to reach the same performance that a model which does not use it. Specifically, the performance of an experiment with 20-patients database and data augmentation is comparable with a one with 30-patients database and no data augmentation. 

In addition, we saw an incremental evolution of the performance metrics when the database increased as well, but between 40 and 47 patients, it was seen stabilization and decreasing in the variance. To put in context, the variance between data augmentation and no data augmentation when 10 patients were 7\% and 16\%, respectively; similarly, for 20 (10,1\% in both), 30 (11\% and 13\%), 40 (9\% and 1\%) and 47 patients (5\% and 4\%) there was a constant decrease in variance, therefore showing the models’ robustness improving. In conclusion, the CNNs performance is a trade-off between data augmentation versus database size, the higher the database’ volume, the higher the performance. Likewise, the more data augmentation the better. This is a far-reaching conclusion, which gives helpful insights for further experimentation with the DMR-IR database or for researchers, which seek to conceive new breast thermography databases. Therefore, this pioneering study could clarify upcoming experimentation with breast thermograms, where there is no information on how big the database should be in order to obtain acceptable performances. 

Finally, we note that the application of this work is centered on demonstrating that CNN-based CAD systems are more viable than the ones based on texture and statistical features because of both robustness and easy implementation. We have reviewed several studies, their techniques and methodologies towards databases of thermal image for a breast cancer diagnosis; nevertheless, it is important to mention some limitations. Firstly, the lack of information (thermal images) limits the generalization that an MLT could reach (the more data the better). Secondly, the physicians and researchers expect to know what the algorithm is computing, but normally the CNN models are recognized as black box MLT; thus, innovative techniques are measuring the CNN’s inside behavior throughout the training process. Further research in this area could clarify some still unanswered questions. Thirdly, the physicians prefer systems that give an image and a probability rather than a merely probability of having cancer; therefore, future work should develop CAD systems that deal with these issues. 

To finish up, this article proposes a novel CNN-based method for breast cancer diagnosis using thermal images. We showed that a well-delimited database split technique is needed in order to reduce the bias and overfitting during the training process. The paper presents the last studies on the DMR-IR database. Experimental results confirm that our database split methodology minimizes the overfitting and bias during training. In addition, this paper conveys the first state-of-the-art benchmark of CNN architectures such as ResNet, SeResNet, VGG16, Inception, InceptionResNetV2 and Xception for the DMR-IR database. Likewise, this study establishes the first CNN hyper-parameters optimization in a thermography database for breast cancer, where the top CNN model achieved a 92\% accuracy, 94\% precision, 91\% sensitivity and 92\% F1-score. We demonstrated that the trade-off between database size and data augmentation techniques are crucial in classification tasks lacking sufficient data such as the one presented in the present study. We have demonstrated that CAD systems for breast cancer diagnosis with thermal images can be valuable and reliable additional tools for physicians, but further research is needed on bigger databases and in multi-class classification problems.

\section*{Disclosure statement}
The authors have stated that they have no conflicts of interest.

\section*{Funding}
This work has been supported by the INTERREG (France - Switzerland) under the SBRA project.

\section*{Acknowledgement}
JZ, ZA, KB, SM and NZ contributed to conception and design. JZ led the data pre-processing and algorithm conception. JZ, ZA, KB and SM contributed to analysis and discussion of the results. JZ contributed with the preparation of the manuscript. JZ, ZA, KB, SM and NZ contributed with to the reviews of the manuscript. All authors read and approved the final manuscript. This work has been supported by the EIPHI Graduate school (contract "ANR-17-EURE-0002").

\bibliographystyle{unsrt}  
\bibliography{references}
\end{document}